\pdfoutput=1
\newcount\Comments  
\documentclass[11pt]{article}

\PassOptionsToPackage{table,xcdraw}{xcolor}
\usepackage[]{ACL2023}
\usepackage{color}
\usepackage{xcolor}
\usepackage{listings}
\usepackage{tcolorbox}\tcbuselibrary{skins}
\usepackage{soul}
\usepackage{siunitx}
\newcounter{Comments}
\newcommand{\kibitz}[2]{%
  \ifnum\value{Comments}=1
    \textcolor{#1}{#2}%
  \fi
}
\definecolor{codegreen}{rgb}{0,0.6,0}
\definecolor{codegray}{rgb}{0.5,0.5,0.5}
\definecolor{backcolour}{HTML}{ECF7FB}
\definecolor{emph}{RGB}{166,88,53}
\definecolor{nightblue}{RGB}{9,49,105}
\definecolor{keywords}{RGB}{207,33,46}
\definecolor{lightpurple}{RGB}{130,81,223}
\newcommand{\fon}[1]{\fontfamily{#1}\selectfont} 
\definecolor{CB_babyblue}{HTML}{A4DDED} 
\tcbset{prompt/.style={
    enhanced,
    size=fbox,
    boxrule=3pt,
    arc=2mm,
    auto outer arc,
    left=10pt,
    right=10pt,
    top=10pt,
    bottom=10pt,
    colback=CB_babyblue!15,
    colframe=CB_babyblue!30,
    coltitle=CB_babyblue!25!black, 
    fontupper=\fon{cmtt}, 
}}
\definecolor{darkgreen}{rgb}{0,0.5,0}
\definecolor{purple}{rgb}{1,0,1}

\usepackage{times}
\usepackage{latexsym}
\usepackage{amsmath}
\usepackage{graphicx}
\usepackage{hyperref}
\usepackage{booktabs}
\usepackage{multirow}
\usepackage{tabularx} 
\usepackage{array} 
\tcbuselibrary{listings}
\lstdefinestyle{mystyle}{
    backgroundcolor=\color{backcolour},   
    commentstyle=\color{codegreen},
    keywordstyle=\color{keywords},
    stringstyle=\color{nightblue},
    basicstyle=\fontsize{8.5}{10}\fon{cmtt},
    breakatwhitespace=true,         
    breaklines=true,                 
    captionpos=b,                    
    keepspaces=true,                 
    numberstyle=\tiny\color{codegray},
    numbersep=2pt,                  
    showspaces=false,                
    showstringspaces=false,
    showtabs=false,                  
    tabsize=2,
    emph={dspy},
    emphstyle={\color{lightpurple}},
    linewidth=1\textwidth,
    frame=tb,    
    xrightmargin=0pt,
    xleftmargin=0.23cm,
    numbers=left,
    aboveskip=0.2cm,
    belowskip=0.1cm,
}

\tcbset{
  pythonbox/.style={
    listing engine=listings,
    colback=white,         
    colframe=black,        
    listing only,
    listing options={style=emnlp},
    sharp corners,
    boxrule=0.5pt,
    top=1mm,bottom=1mm,left=1mm,right=1mm,
    title={\textbf{LLM-as-a-Judge: Comparing two English translations}},
  }
}

\usepackage[T1]{fontenc}
\usepackage[T5]{fontenc}

\usepackage[utf8]{inputenc}
\usepackage{cleveref}
\crefname{section}{§}{§§}

\usepackage{microtype}
\usepackage{array}
\usepackage{calc}
\usepackage{enumitem}

\usepackage{inconsolata}
\newcommand{\langcode}[1]{\textbf{\texttt{#1}}}

\title{VietMix: A Naturally-Occurring Parallel Corpus and Augmentation Framework for Vietnamese-English Code-Mixed Machine Translation}

\author{
    Hieu Tran$^{*,1}$, Phuong-Anh Nguyen-Le$^{*,1}$, Huy Nghiem$^{1}$, Quang-Nhan Nguyen$^{2}$, \\
    \textbf{Wei Ai}$^{1}$, \textbf{Marine Carpuat}$^{1}$\\
    $^{1}$ University of Maryland, College Park, $^{2}$ Harvard University\\
    $^{1}$\{hieutt, nlpa, nghiemh, aiwei, marine\}@umd.edu\\
    $^{2}$quangnhan\_nguyen@fas.harvard.edu \\
}

\begin{document}
\maketitle
\renewcommand{\thefootnote}{$*$}
\footnotetext{These authors contributed equally to this work.}
\renewcommand{\thefootnote}{\arabic{footnote}}
\setcounter{footnote}{0}

\begin{abstract}

Machine translation (MT) systems universally degrade when faced with code-mixed text. This problem is more acute for low-resource languages that lack dedicated parallel corpora. This work directly addresses this gap for Vietnamese-English, a language context characterized by challenges including orthographic ambiguity and the frequent omission of diacritics in informal text. We introduce \textsc{VietMix}, the first expert-translated, naturally occurring parallel corpus of Vietnamese-English code-mixed text. We establish \textsc{VietMix}'s utility by developing a data augmentation pipeline that leverages iterative fine-tuning and targeted filtering. Experiments show that models augmented with our data outperform strong back-translation baselines by up to +3.5 xCOMET points and improve zero-shot models by up to +11.9 points. Our work delivers a foundational resource for a challenging language pair and provides a validated, transferable framework for building and augmenting corpora in other low-resource settings. 

\end{abstract}

\section{Introduction}\label{sec:intro}

 The proliferation of code-mixing, the blending of multiple languages within a single utterance \cite{poplack1980sometimes}, is dominant in global digital communication. While this reflects the realities of multilingualism, it severely degrades the performance of state-of-the-art machine translation (MT) systems, especially those overwhelmingly trained on non-code-mixed formal text \cite{zhang-etal-2023-multilingual}. In low-resource language pairs, the absence of large-scale code-mixed parallel corpora creates a cycle of poor model performance and data scarcity. Benchmark datasets have emerged for medium-resource pairs, e.g., \citet{sheth2025comi}, but low-resource communities remain underserved. 
 
 Throughout this paper, we adopt the sociolinguistic term "code-mixing" to refer specifically to intra-sentential blending of linguistic units \cite{bokamba1989there}. While NLP works often use "code-mixing" as an umbrella term to cover all forms of language alternation \cite{dogruoz-etal-2021-survey, winata-etal-2023-decades}, our choice of "code-mixing" more precisely describes the hybridization phenomena that characterize our corpus.  

 This paper focuses on Vietnamese-English, where a confluence of unique linguistic properties amplifies the code-mixing challenge. First, the two share significant orthographic ambiguity. For instance, the token \colorbox{gray!20}{ban} is the English word meaning ("prohibition") but corresponds to at least five distinct Vietnamese words with different tones (bạn (friend), bản (village), bàn (table), bán (to sell), ban (to grant)), rendering context-free language identification nearly impossible. Second, informal digital text frequently omits the diacritics essential for disambiguating meaning in Vietnamese. Third, the prevalence of English-origin proper nouns and named entities further confounds this task of identifying true code-mixing. Automatic data curation, therefore, becomes exceptionally difficult to achieve and, until now, has hindered the development of a dedicated natural corpus.  

Addressing this, we introduce \textsc{VietMix}, the first expert-translated parallel corpus of naturally occurring Vietnamese-English code-mixing (\cref{sec:vietmix}). Our core contribution lies in both this resource and the validated methodology for its creation. \textsc{VietMix} contains over 10K code-mixed samples collected from Threads.com, with a held-out test set of 1K sentences translated and adjudicated by linguistic experts to serve as a gold-standard evaluation. 

 Accompanying \textsc{VietMix}, we developed a three-stage data augmentation pipeline to establish strong baselines and provide a transferable framework for the community (\cref{sec:augmentation}). Critically, we show that augmenting models with our filtered, synthetic data provides consistent and significant gains over zero-shot models  by up to $+11.9$ xCOMET points and strong back-translation baselines by up to $+3.5$ points. This proves the value of our quality-focused curation approach. Experimental validation across multiple model architectures (\cref{sec:experiments}) confirms that this methodology provides more faithful translations, especially in capturing pragmatic and idiomatic nuances of code-mixed language (\cref{sec:results}).

  We release the held-out test set under a Data Use Agreement (non-commercial, research-only). Code and preprocessing recipes are publicly accessible.\footnote{https://github.com/umd-vietnlp/code-switching}

\begin{figure*}[htbp]
    \centering
    \includegraphics[width=\textwidth]{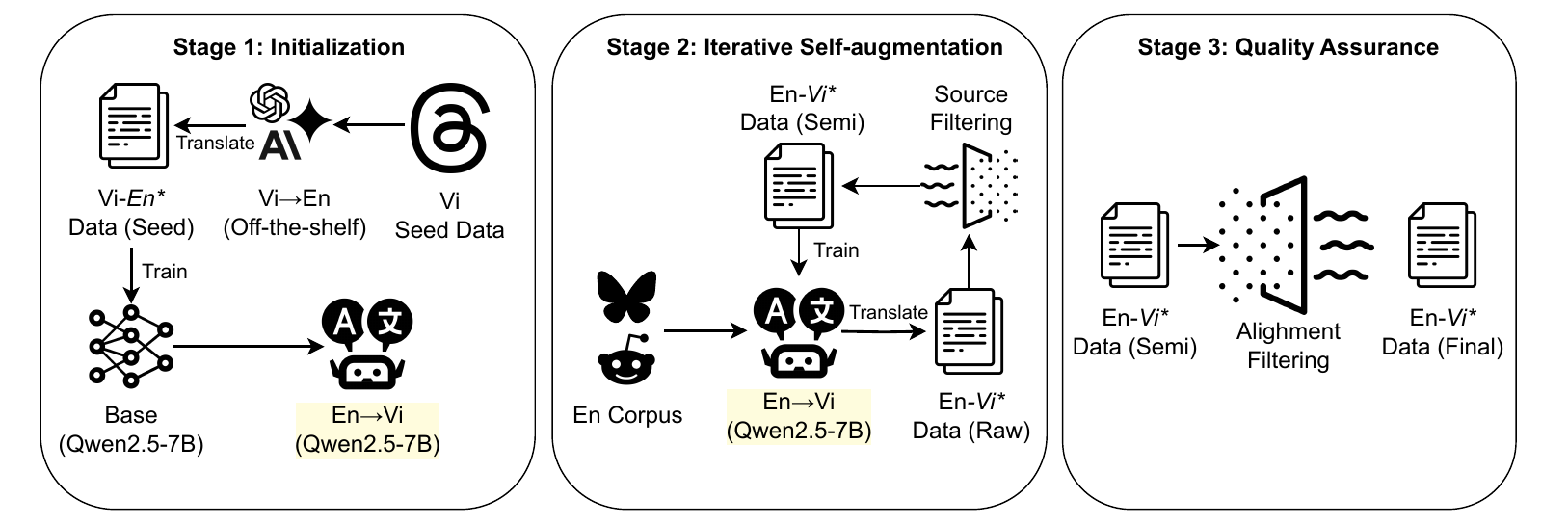}
    \caption{Overview of our three-stage iterative data augmentation pipeline converting naturally occurring code-mixed \langcode{Vi} texts into synthetic parallel \langcode{Vi-En} corpora. \textbf{Stage 1} focuses on seed data generation from naturally occurring \langcode{Vi} text and base model (\langcode{En}$\rightarrow$\langcode{Vi} translation) bootstrapping. \textbf{Stage 2} involves iterative synthetic \langcode{Vi} generation from \langcode{En} corpora and progressive fine-tuning with filtering. \textbf{Stage 3} implements final quality assurance to produce the augmented parallel dataset. \textit{Italicized language code} denotes synthetic/generated data. 
    } 
    \label{fig:pipeline}
\end{figure*}

\section{Related Work}\label{sec:lit}

Our work intersects three research areas: code-mixed MT, resource creation for low-resource languages, and synthetic data generation.

\paragraph{Code-mixed MT.}

The challenge that mixed-language text poses to NLP systems is well-documented, with initial works focusing on core tasks like LID and part-of-speech tagging \cite{solorio-liu-2008-part, li-fung-2012-code}. Centralized benchmarks like GLUECoS \cite{khanuja-etal-2020-gluecos} and LinCE \cite{aguilar-etal-2020-lince} enabled systematic evaluation of model degradation on code-mixed inputs. These efforts have shown that even large multilingual models struggle to transfer non-code-mixed capabilities to code-mixed scenarios. 

Within this landscape, MT presents distinct challenges, primarily in the scarcity of parallel data. Consequently, research has concentrated on a few relatively high-resource language pairs, with Hindi-English being the most prominent with annotated datasets \cite{sheth2025comi} and specialized augmentation techniques \cite{kartik-etal-2024-synthetic}. While work has expanded to low-resource pairs like Kazakh-Russian \cite{borisov-etal-2025-low} and dialects of Mandarin \cite{lu-etal-2022-exploring}, these efforts remain more limited in scope and do not address unique challenges in Southeast Asian tonal languages. Our work addresses this linguistic and geographic gap.

\paragraph{Parallel Corpora for Vietnamese MT.}

 The ecosystem of NLP resources for Vietnamese has matured significantly from foundational work on tokenization and syntactic parsing \cite{vu-etal-2003-towards, nguyen-etal-2006-vietnamese} to the comprehensive VnCoreNLP toolkit \cite{vu-etal-2018-vncorenlp}.  In MT, large-scale multi-domain corpora like PhoMT \cite{doan-etal-2021-phomt} and MTet \cite{ngo2022mtet} have been instrumental in advancing translation quality for non-code-mixed Vietnamese-English. Recent efforts focused on domain specialization, namely, medical texts \cite{vo-etal-2024-improving} and other lanuguage pairs like Japanese-Vietnamese \cite{do2021mining} or Korean-Vietnamese \cite{nguyen2019korean}.

Despite the pervasiveness of code-mixing in Vietnamese social media, prior work on Vietnamese code-mixing has been limited to small-scale and non-parallel speech corpora \cite{nguyen-bryant-2020-canvec} or monolingual text mining \cite{nguyen-etal-2024-vilexnorm, hoang-etal-2023-vihos}. \textsc{VietMix} is the first resource to fill this gap and provides a gold-standard, expert-translated benchmark to catalyze research on this phenomenon. 





\paragraph{Synthetic Data for Low-Resource MT.}




Given high costs in creating parallel corpora, researchers have long turned to monolingual and comparable web corpora as a source of training data. Early statistical MT methods mined noisy parallels or lexicons to bootstrap low-fidelity translation systems before iteratively adding comparable monolingual corpora for retraining \cite{abdulrauf2009exploiting, do-etal-2010-fully}. More recent neural methods have refined this process using multilingual embeddings and self-supervised retraining to extract parallel data with minimal supervision \cite{keung2020unsupervised, shi2022obtaining}.  

LLMs introduce a new paradigm for synthetic data generation as they can output vast amounts of parallel data through both forward- and back-translation \cite{nguyen-etal-2024-democratizing, lim2024mufu}, with back-translation often yielding higher adequacy improvements \cite{moslem-etal-2022-domain}. However, naive back-translation can distort code-mixing patterns or "correct" them into non-code-mixed text and compound errors at language-switching boundaries. This failure highlights the need for more robust, quality-aware augmentation protocols. Our work builds on this precedent by proposing a forward-translation pipeline coupled with a multi-stage filtering mechanism, combining heuristic constraints and a neural quality classifier, to ensure syntactic plausibility and pragmatic appropriateness of the generated synthetic code-mixed.

\section{\textsc{VietMix} Curation and Annotation}\label{sec:vietmix}
This section details the multi-stage process for creating \textsc{VietMix}. Our methodology emphasizes rigorous data filtering, expert human translation, and transparent translation quality control. 

\subsection{Data Sourcing}\label{data:collection}
\textsc{VietMix} consists of public posts from the social media Threads.com between October 2024 and February 2025. Adhering to strict data minimization and ethical research principles, we scraped publicly accessible content without user authentication using public APIs.

\subsection{Data Preprocessing and Filtering}

To ensure the quality of the corpus, the raw data underwent a comprehensive filtering pipeline to isolate relevant code-mixed content. We first applied several heuristic filters, removing posts that were overly short ($\leq$10 words), contained links, consisted solely of emojis or special symbols.

Following this initial cleanup, a crucial step was to perform sentence-level Language Identification (LID) to accurately identify true code-mixing. This is a non-trivial task, as off-the-shelf LID tools are often unreliable for Vi-En due to high orthographic overlap (e.g., the word "ban") and the common omission of diacritics. To address this, we developed a robust two-stage LID process:
\begin{enumerate}
    \setlength{\itemsep}{0pt} \setlength{\parskip}{0pt} \setlength{\parsep}{0pt}
    \item \textbf{Initial Filtering.} We used the \texttt{lingua} library for a high-recall pass to identify all Vietnamese posts containing English tokens.
    \item \textbf{Ensemble LLM Validation.} This initial subset was then processed by a majority-vote ensemble of three LLMs (GPT-4o~\cite{openai2024gpt4technicalreport}, Claude-3.5~\cite{anthropic2024claude3}, and Gemini-1.5~\cite{geminiteam2024gemini15unlockingmultimodal}) to accurately classify samples, resolving ambiguities in orthography and named entities.
\end{enumerate}

We validated this pipeline on a manually annotated 400-sample set, where it achieved a $87.98\%$ F1-score and $89.75\%$ accuracy, significantly outperforming baseline methods (see Appendix~\ref{app:lid}).

The final stage of our filtering pipeline is the removal of personally identifiable information (PII) to ensure user privacy. All data undergoes an automated process to detect and redact sensitive spans like phone numbers, email addresses, and similar artifacts. To guarantee the integrity of our benchmark, the test set undergoes an additional, rigorous manual review to identify and handle any complex edge cases missed by the automated system. Across all splits, posts containing PII that cannot be cleanly redacted are discarded entirely.

\subsubsection{Final Corpus Statistics and Splits}
Applying our validated pipeline to the sourced data yields the final \textsc{VietMix} corpus of 10{,}254 high-confidence code-mixed posts. We then perform a random split into a training set (8{,}791 posts; seed data for synthetic generation), a development set (462 posts; \emph{source-only} with no target references, used for model selection and ablations via reference-free models), and a held-out test set (1{,}002 posts; reserved for expert translation). Table~\ref{tab:data-stats} reports corpus statistics, including the Code-Mixing Index (CMI; \citealt{srivastava-singh-2021-challenges}) and Switching Point Frequency (SPF, \citealt{pratapa-etal-2018-language}).

\begin{table}[t]
\centering
\small
\begin{tabular}{lcccc}
\toprule
\multirow{2}{*}{\textbf{Statistic}}
  & \multicolumn{3}{c}{\textbf{Seed}}
  & \multirow{2}{*}{\textbf{Aug.}} \\ 
\cmidrule(lr){2-4}
  & Train   & Dev     & Test    \\ 
\midrule
CMI                & 21.66   & 21.36   & 20.66   & 18.20    \\
SPF                & 0.49    & 0.49    & 0.49    & 0.48     \\
\# tokens (src)    & 372,415& 17,179 & 31,801 & 3,460,247 \\
\# tokens (tgt)    & 369,429&  0       & 31,952 & 3,347,241 \\
\# samples         & 8,790   & 462     & 1,002   & 81,768   \\
\bottomrule
\end{tabular}
\caption{Corpus statistics for the naturally occurring Vi-En code-mixed seed data splits (Train, Dev, Test) and the synthetically generated augmented (Aug.) data. Metrics include Code-Mixing Index (CMI), Switching Point Frequency (SPF), token counts for source (src) and target (tgt), and number of samples. The `Test' set of the seed data forms the basis of the VietMix corpus.}
\label{tab:data-stats}
\end{table}

\subsection{Expert Translation of the Gold-Standard Test Set}\label{data:test_set}
To create a gold-standard benchmark, the 1,002 samples in the held-out test set were translated into English. This was performed by two professional translators with expertise in social media language, following a rigorous quality control process:

\begin{itemize}
    \setlength{\itemsep}{0pt}
    \setlength{\parskip}{0pt}
    \setlength{\parsep}{0pt}
    \item \textbf{Phase 1 (Alignment)}: Translators first independently translated 70 samples deemed as most challenging. They then cross-evaluated each other's work, rating on an adapted version of the MQM-Core framework\footnote{\url{https://themqm.org/the-mqm-typology/}}, which focuses on six dimensions, including accuracy, style, and locale conventions (Appendix~\ref{sec: MQM}). Disagreements were mediated by a Vietnamese language instructor to establish clear conventions for handling challenging cases (e.g., puns, cultural references, neologisms). 
    \item \textbf{Phase 2 (Translation \& Validation)}: Translators then completed the remaining of the data. All translations underwent a final verification pass by our language expert to ensure adherence to established guidelines and refine for maximum semantic fidelity.  
\end{itemize}

This process ensures \textsc{VietMix} becomes a high-quality, ecologically valid benchmark suitable for evaluations of code-mixed MT systems.

\section{Quality-Aware Data Augmentation}\label{sec:augmentation}

\begin{figure}[t]
    \centering
    \includegraphics[width=\columnwidth]{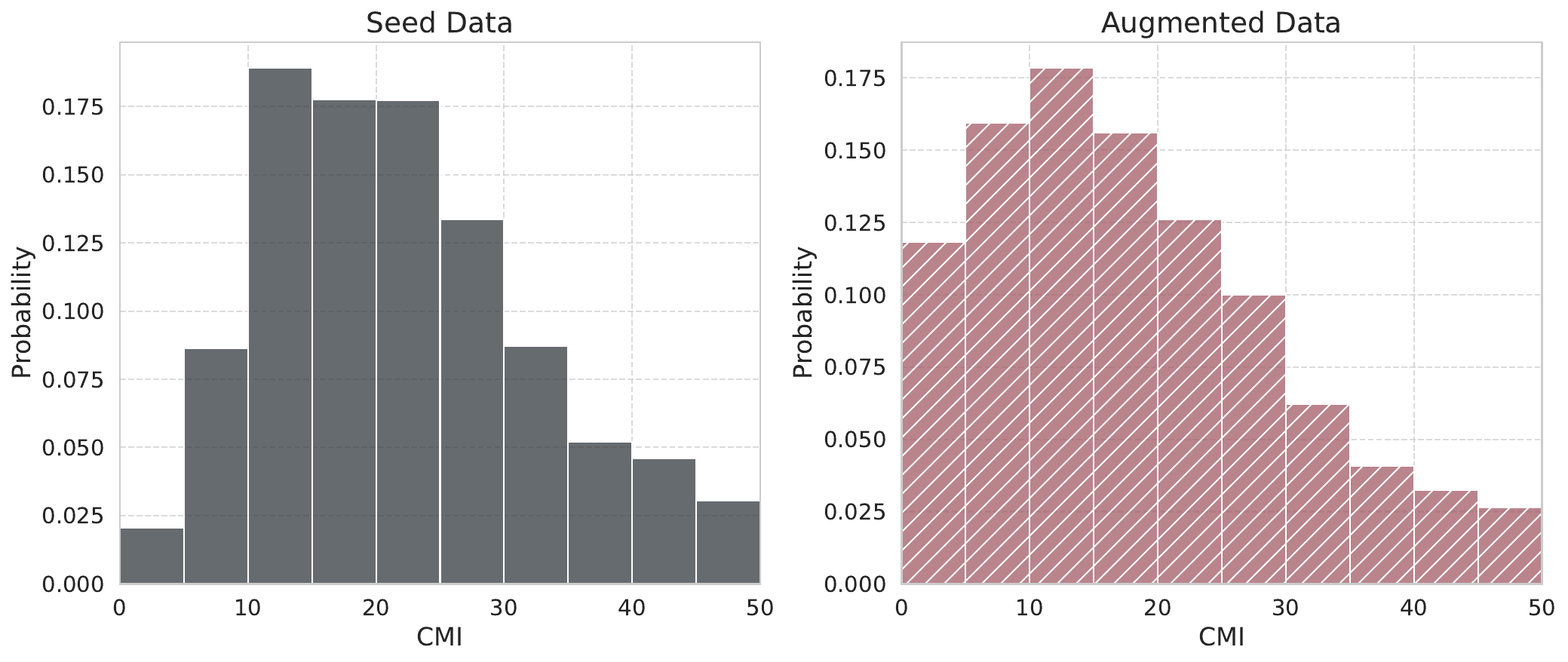}
    \caption{Comparative distribution of the Code-Mixing Index (CMI) in the naturally occurring VietMix seed corpus (Left/Real) versus the synthetically generated augmented corpus (Right/Augmented). The CMI quantifies the extent of language mixing within text samples.}
    \label{fig:cmi_distribution}
\end{figure}

The \textsc{VietMix} corpus provides a high-quality foundation, but its size is insufficient for training robust systems. Standard augmentation techniques like back-translation often fail on code-mixed text, correct it into a non-code-mixed form, or produce unnatural translations at language boundaries. Addressing this, we developed a three-stage data augmentation pipeline designed to generate a large-scale, high-quality synthetic parallel Vietnamese-English code-mixing corpus. Our pipeline's architecture, visualized in Figure~\ref{fig:pipeline}, prioritizes quality through iterative retraining and filtering. 

\paragraph{Stage 1: Seed Data Generation and Base Model Bootstrapping.} The pipeline begins with naturally-occurring code-mixed Vietnamese (\langcode{Vi}) from our training split (n = 8,791). 
\begin{itemize}
    \setlength{\itemsep}{0pt}
    \setlength{\parskip}{0pt}
    \setlength{\parsep}{0pt}
    \item \textbf{Initial Translation.} We employ GPT-4o to generate initial English translations (\langcode{En}) for this \langcode{Vi} sample. This creates the natural \langcode{Vi} $\leftrightarrow$ synthetic \langcode{En} seed parallel data.
    \item \textbf{Base Model Fine-tuning.} We then fine-tune a pre-trained multilingual model, \texttt{Qwen2.5-7B}, on this seed data for the \langcode{En}$\rightarrow$\langcode{Vi} translation task. \texttt{Qwen2.5-7B} has strong documented performance on East Asian languages, including Vietnamese \cite{qwen2.5}. This model is now "bootstrapped" with a foundational understanding of \langcode{Vi} and can serve as the core engine for subsequent augmentation. 
\end{itemize}

\paragraph{Stage 2: Iterative Synthetic Data Generation.} This stage generates large volumes of synthetic \langcode{Vi} that mirror real \langcode{Vi} in style and domain. 

\begin{itemize}
    \setlength{\itemsep}{0pt}
    \setlength{\parskip}{0pt}
    \setlength{\parsep}{0pt}
    \item \textbf{Source Data Selection.} We select large monolingual English corpora from sources stylistically similar to our seed data (Bluesky\footnote{\url{https://huggingface.co/datasets/Roronotalt/bluesky}} and Reddit\footnote{\url{https://huggingface.co/datasets/sentence-transformers/reddit-title-body}}) to minimize domain mismatch and ensure the generated text reflects the informal, conversational register of social media.
    \item \textbf{Iterative Translation and Retraining.} Using the bootstrapped \langcode{En}$\rightarrow$\langcode{Vi} model from Stage 1 to translate the English source text, we produce a large corpus of synthetic \langcode{Vi}. Iteration is key to this stage. After generating an initial batch of \langcode{Vi}, we apply quality filters as described in Appendix~\ref{appendix:filtering}, combine the filtered synthetic data with our original seed, and retrain the \langcode{En}$\rightarrow$\langcode{Vi} translation model. Iteration nudges the model to progressively generate more natural code-mixed patterns. 
\end{itemize}



\paragraph{Stage 3: Final Quality Assurance.} Several cycles of iterative generation and filtering aggregate to a final corpus of over 200K pairs. As a final quality check, we use a reference-free quality estimation metric, xCOMET \cite{guerreiro-etal-2024-xcomet}, to score the semantic alignment of each \langcode{En}-\langcode{Vi} pair. Through manual inspection, we established 0.9 as a threshold and discarded all pairs below this score to prune examples with significant meaning shifts.  

Our pipeline yielded a final augmented corpus of over 80K parallel examples. We use this corpus in our experiments to measure the impact of quality-aware synthetic data generation. The CMI distribution of this final corpus compared against the original seed is reported in Figure \ref{fig:cmi_distribution}. 


\section{Experimental Validation}\label{sec:experiments}

We conduct a series of experiments to empirically validate the utility of our contributions. Our main objective is to determine the extent fine-tuning on the \textsc{VietMix} corpus and our quality-aware augmented data improve code-mixed MT performance.


\subsection{Experimental Setup}
We evaluate three diverse open-source multilingual backbones to reduce architecture-specific effects: \texttt{GemmaX2-28-9B}, a sparse Mixture-of-Experts model specialized for multilingual translation with Vietnamese in pre-training \cite{cui-etal-2025-multilingual}; \texttt{Qwen2.5-7B}, a dense model with strong performance in East Asian languages \cite{qwen2.5}; and \texttt{Llama3.1-8B}, a general-purpose dense model with predominantly English pre-training to serve as contrast to multilingual models \cite{grattafiori2024llama3herdmodels}. Unless otherwise noted, we fine-tune the \emph{base} variants using Low-Rank Adaptation (LoRA). Full hyperparameters are in Appendix~\ref{appendix:experimental_details}.

We assess each backbone under four training conditions to isolate the contribution of our data:
\begin{description}[leftmargin=0pt,labelsep=6pt,itemsep=2pt,parsep=0pt,topsep=2pt]
  \item[\textbf{Instruction-tuned (zero-shot).}]
  Off-the-shelf instruction variants without additional fine-tuning; used as lower-bound baselines.

  \item[\textbf{Back-Translation (BT).}]
  Trained on 80K{+} synthetic pairs via standard back-translation, using the English side of our augmented corpus (\cref{sec:augmentation}) and regenerating Vietnamese with an \langcode{En}$\rightarrow$\langcode{Vi} model fine-tuned on the \textsc{VietMix} seed data.

  \item[\textbf{Seed-only}.]
  Base variants fine-tuned exclusively on the \textsc{VietMix} seed set (8{,}791 pairs).

  \item[\textbf{Augmented (Ours).}]
  Base variants fine-tuned on our quality-filtered synthetic corpus (80K{+} pairs).
\end{description}



\subsection{Evaluation Protocol}

Our primary semantic metrics are reference-based xCOMET\footnote{\url{https://huggingface.co/Unbabel/XCOMET-XXL}} and reference-free COMETKiwi\footnote{\url{https://huggingface.co/Unbabel/wmt23-cometkiwi-da-xxl}}. Both were chosen for their support for Vietnamese and their demonstrated strong correlation with human judgments in recent WMT evaluations across multiple domains, especially for domains like social media chat \citep{freitag-etal-2022-results,rei-etal-2023-scaling,guerreiro-etal-2024-xcomet}. We avoid BLEU as a headline metric as prior work shows that typical test-set references exhibit limited diversity and tend to concentrate around translationese, which depresses BLEU and weakens its alignment with human preferences \citep{freitag-etal-2020-bleu, wu-etal-2024-transagents}. For broader comparability with other studies, we also report scores for BLEU in Appendix~\ref{appendix:experimental_details}.

To complement the automatic scores, we employ an LLM-as-a-Judge framework using GPT-4.1 for pairwise preference judgments between outputs from our \emph{Seed-only} and \emph{Augmented} models. This provides a qualitative validation of the improvements indicated by the metrics \citep{kocmi-federmann-2023-large}. For structured and reliable outputs, prompts were managed using DSPy \cite{khattab2024dspy}, with full specifications reported in Appendix~\ref{appendix:judge_prompts}.

To ensure the reliability of our findings, we employ paired permutation tests to determine if the observed improvements across system pairs are statistically significant.\footnote{\url{https://docs.scipy.org/doc/scipy/reference/generated/scipy.stats.permutation_test.html}}

\definecolor{DeltaBlue}{RGB}{198,219,239}   
\definecolor{DeltaGreen}{RGB}{198,239,206}  
\definecolor{DeltaYellow}{RGB}{255,242,204} 

\begin{table}[t]
  \scriptsize
  \setlength{\tabcolsep}{3pt}
  \renewcommand{\arraystretch}{1}
  \centering
  \resizebox{\columnwidth}{!}{%
    \begin{tabular}{llcc|cc}
      \toprule
      \multicolumn{2}{c}{\textbf{Model}} &
      \textbf{xCOMET} & \textbf{$\Delta$} &
      \textbf{COMETKiwi} & \textbf{$\Delta$} \\
      \midrule
      \multicolumn{6}{l}{\textbf{Instruction models}} \\
       & \texttt{Qwen2.5-7B-Inst}     & 77.72 & -- & 67.16 & -- \\
       & \texttt{Llama3.1-8B-Inst}    & 66.96 & -- & 57.96 & -- \\
       & \texttt{GemmaX2-28-9B-v0.1}  & 77.83 & -- & 65.85 & -- \\
      \midrule
      \multicolumn{6}{l}{\textbf{Back-Translation (BT)}} \\
       & \texttt{Qwen2.5-7B}          & 78.27 & -- & 65.94 & -- \\
       & \texttt{Llama3.1-8B}         & 77.57 & -- & 64.02 & -- \\
       & \texttt{GemmaX-9B}           & 78.26 & -- & 67.51 & -- \\
      \midrule
      \multicolumn{6}{l}{\textbf{Ours}} \\
      \multicolumn{6}{l}{\quad\textbf{with Seed only}} \\
       & \texttt{Qwen2.5-7B}          & 78.74 & -- & 69.72 & -- \\
       & \texttt{Llama3.1-8B}         & 78.19 & -- & 68.79 & -- \\
       & \texttt{GemmaX-9B}           & 80.31 & -- & 71.11 & -- \\
      \addlinespace
      \multicolumn{6}{l}{\quad\textbf{with Augmented}} \\
       & \texttt{Qwen2.5-7B}          & 79.28 & \cellcolor{DeltaGreen}{$+\!0.54$} & 69.99 & \cellcolor{DeltaYellow}{$+\!0.26$} \\
       & \texttt{Llama3.1-8B}         & 78.90 & \cellcolor{DeltaGreen}{$+\!0.71$} & 69.44 & \cellcolor{DeltaGreen}{$+\!0.65$} \\
       & \texttt{GemmaX-9B}           & \textbf{81.73} & \cellcolor{DeltaBlue}{\textbf{$+\!1.41$}} & \textbf{71.60} & \cellcolor{DeltaGreen}{$+\!0.49$} \\
      \bottomrule
    \end{tabular}%
  }
  \caption{Performance on \langcode{Vi$\rightarrow$En} (VietMix test). We report reference-based xCOMET and reference-free COMETKiwi for instruction-tuned baselines, back-translation (BT), our models fine-tuned only on the seed data, and our models fine-tuned with the augmented dataset. Higher is better. The $\Delta$ columns show \emph{Augmented} minus \emph{Seed-only}. Colored boxes highlight performance differences compared to \emph{Seed-only}: blue = strong ($p{<}.001$), green = significant ($p{<}.05$), yellow = non-significant ($p{>}.05$).}
  \label{tab:model_performance}
\end{table}

\section{Results} \label{sec:results}

We analyze empirical results of our experiments using quantitative metrics, followed by a qualitative error analysis to provide a holistic view.

\subsection{Main Quantitative Results}

 Table~\ref{tab:model_performance} presents our primary results. All models improve significantly when fine-tuned on the \textsc{VietMix} seed corpus compared to their zero-shot performance, confirming that even a small amount of high-quality, in-domain data is crucial. Building on this, our quality-aware augmentation consistently and substantially outperforms the BT baseline across all models. For instance, \texttt{GemmaX-9B} improves to $81.73$/$71.60$ ($+3.47$/$+4.09$ over BT), with \texttt{Llama3.1-8B} and \texttt{Qwen2.5-7B} showing similar strong gains. This validates that a principled, quality-centered strategy is superior to naive synthetic data generation.

Our augmentation also provides a consistent boost over the \emph{Seed-only} models, though the gains vary in an insightful way. While \texttt{GemmaX-9B} ($+1.41$/$+0.49$) and \texttt{Llama3.1-8B} ($+0.71$/$+0.65$) benefit significantly, the COMETKiwi gain for \texttt{Qwen2.5-7B} is minimal ($+0.26$). This is expected, as \texttt{Qwen2.5-7B} was the generator for the augmented data. Fine-tuning a model on its own outputs offers little new signal and can reinforce existing biases. In contrast, the other models benefit from the augmented pairs as teacher-style supervision from a different model, leading to more effective knowledge transfer. Overall, the largest incremental gains appear with \texttt{GemmaX-9B}, while all backbones benefit from our quality-aware augmentation.


 \subsection{Validating the Evaluation Protocol}
Automated MT metrics can be unreliable on code-mixed text. We therefore validated our protocol for \langcode{Vi}-\langcode{En} code-mixing. We selected xCOMET and COMETKiwi because they (i) support Vietnamese and (ii) are widely used in recent MT work
\cite{xu2024contrastivepreferenceoptimizationpushing, cui-etal-2025-multilingual}, while acknowledging that their calibration on \langcode{Vi}-\langcode{En} code-mixed inputs is not well established.

We conducted a focused agreement study between COMETKiwi, xCOMET, and GPT-4.1 acting as a pseudo-human judge (LLM-as-a-judge), which prior work reports can reach $>$80\% agreement with human judgments \citep{zheng2023judgingllmasajudgemtbenchchatbot}. Concretely, we compared GPT-4.1 pairwise preferences between \emph{Seed-only} and \emph{Augmented} system outputs against the “wins” predicted by each COMET metric on the same pairs. A metric declared a win when one translation’s score exceeded the other by more than 2\%; otherwise we treated the case as a tie and excluded it from agreement calculations.\footnote{We adopt a 2\% margin to reduce spurious wins due to small score fluctuations.}

\begin{table}[t]
\centering
\small
\begin{tabular}{lrr}
\toprule
\textbf{Metric} & \textbf{\# Samples} & \textbf{Agreement w/ GPT-4.1} \\
\midrule
COMETKiwi & 493 & 80.1\% \\
xCOMET    & 555 & 81.3\% \\
\bottomrule
\end{tabular}
\caption{Agreement between COMET metrics and GPT-4.1 pairwise preferences on En–Vi code-mixed test items.}
\label{tab:metric-agreement}
\end{table}

As shown in Table~\ref{tab:metric-agreement}, both metrics correlate strongly with the LLM judge's preferences on our code-mixed test set. With agreement rates of $81.3\%$ for xCOMET and $80.1\%$ for COMETKiwi, our study provides considerable confidence that these metrics are reliable proxies for human judgment in this domain. Furthermore, the LLM judge preferred outputs from the \emph{Augmented} model in over $55\%$ of non-tie cases, supporting our earlier evidence of perceptually improved translation quality. We therefore report COMET scores as primary automatic metrics, while recommending continued use of human or LLM-as-a-judge audits for calibration on code-mixed evaluation.

\begin{figure}[t]
    \centering
    \includegraphics[width=\columnwidth]{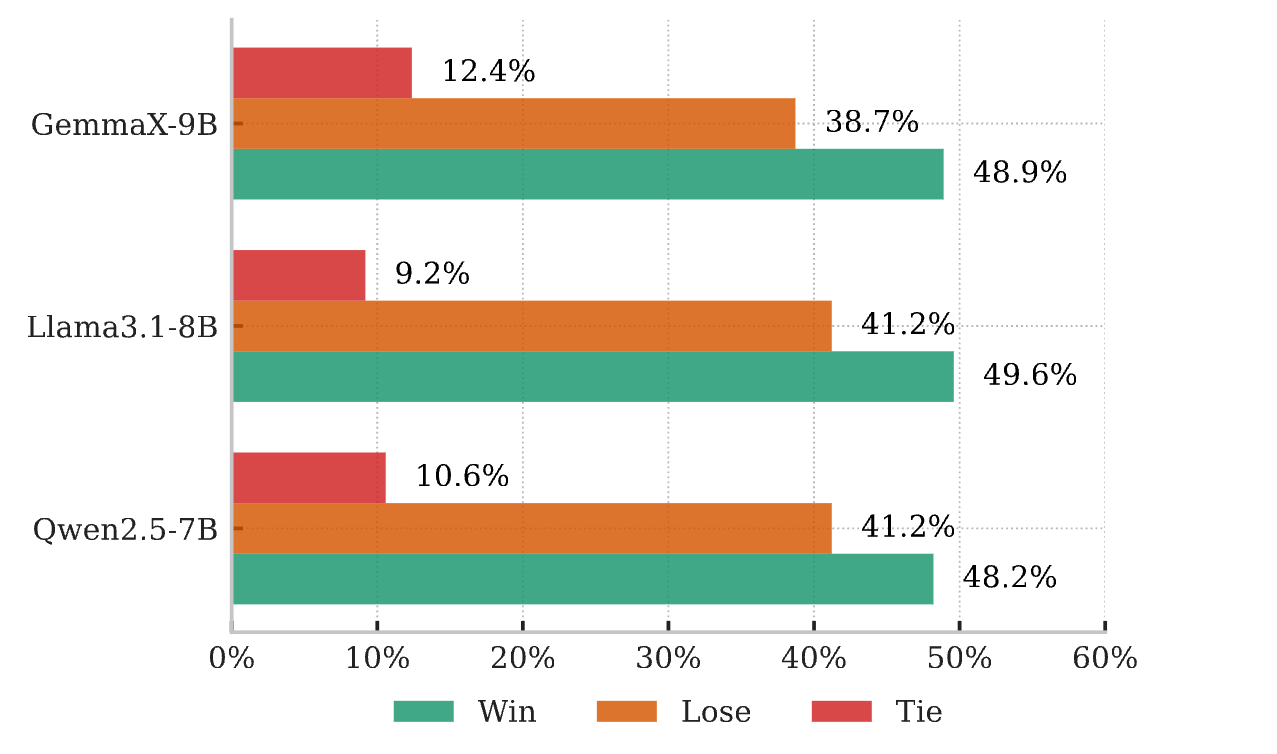}
    \caption{LLM-as-a-Judge assessment comparing \langcode{En} translations from models fine-tuned on augmented data vs. seed data only. Bars represent the percentage of judgments where GPT-4.1 favored the augmented model's translation (Win), the seed-only model's translation (Loss for augmented), or found them equivalent (Tie) across three base models \texttt{(GemmaX2, Llama3.1, Qwen2.5)}. Semantic faithfulness against expert VietMix references was the primary criterion.
    }
    \label{fig:llmjudge_results}
\end{figure}

\subsection{Qualitative Error Analysis}
To understand \emph{how} our approach improves translation beyond aggregate scores, we conducted a manual error analysis over representative test items. We observe consistent gains from the augmentation pipeline in four areas that are central to natural code-mixed communication: pragmatic intent around proper nouns, idiomatic phrasing, transitions at switch boundaries, and acronym handling with contextual disambiguation.

\textbf{Pragmatics near proper nouns.}
Exposure to monolingual \langcode{En} helps models preserve speaker intent when English entities anchor the clause. For “Threads cho lên xu hướng…”, the \emph{Seed-only} output “\textcolor{red}{Threads for trending}…” misses the call-to-action, whereas the \emph{Augmented} output “\textcolor{teal}{Let’s get Threads trending}…” captures the pragmatic force by dropping a literal filler and adopting an imperative framing.

\textbf{Idiomatic phrasing.}
Augmented models learn more faithful colloquial paraphrases, reducing literal calques. For “Loewe puzzle… không bao giờ sợ lỗi thời”, \emph{Seed-only} yields “\textcolor{red}{never afraid of being outdated}”, while \emph{Augmented} produces “\textcolor{teal}{never go out of style}”, a conventional idiom that better matches intent and register.

\textbf{Smoother switch boundaries.}
At complex code-mixed junctures, \emph{Seed-only} systems tend to mash word-for-word fragments across languages (e.g., “\textcolor{red}{apply where pass that}”), whereas \emph{Augmented} systems produce coherent clauses that respect switch-sensitive structure (e.g., “\textcolor{teal}{you will pass everywhere}”). This suggests improved modeling of clause-level alignment rather than token-level overlap.

\textbf{Acronyms and contextual disambiguation.}
Augmented models also resolve acronym-induced ambiguities and country/nationality conflations. In “Du học sinh Mỹ… SSN/ITIN…”, \emph{Seed-only} misreads both actor and identifiers (“\textcolor{red}{American international students}… \textcolor{red}{No social SSN/ITIN}”), while \emph{Augmented} correctly renders “\textcolor{teal}{international students in the US}… \textcolor{teal}{no social security number (SSN/ITIN)}”, leveraging redundant cues in the source to bridge concepts across languages.

\begin{table}[t]
  \centering
  \small
  \begin{tabular}{lcc}
    \toprule
    \textbf{Model}                   & \textbf{w/ Ref} & \textbf{w/o Ref} \\
    \midrule
    Llama3.1-8B (Augmented)         & 78.90           & 80.87            \\
    Llama3.1-8B (9K Seed)           & 78.19           & 80.34            \\
    Llama3.1-8B-Inst                & 66.96           & 67.65            \\
    \bottomrule
  \end{tabular}
  \caption{Comparison of xCOMET scores for \texttt{Llama3.1-8B} translations: the reference-based variant (w/ Ref) versus the reference-free variant (w/o Ref)}
  \label{tab:thread-xcomet}
\end{table}

\begin{table*}[t]
\centering
\small
\begin{tabular}{p{3.5cm} p{3.5cm} p{3.5cm} p{1.5cm} p{2.0cm}}
\toprule
\textbf{Source} & \textbf{Model Hypothesis} & \textbf{Reference} & \textbf{w/ Ref} & \textbf{w/o Ref} \\
\midrule
Khám 120 brands xong vẫn chưa hiểu tại sao chiếc áo bà ba mà bà tư bả mặc & After inspecting 120 brands, I still don't understand why the \textcolor{red}{shirt grandma wears is like that}. & After an inspection of 120 brands, I still don't get why the \textcolor{teal}{fourth lady is wearing the third's shirt}. & 71.43 & 91.31 \\
\midrule
Bé Ánh zia tới q2 rồi muzik mai trói bé zia nhà thiu & Little Ánh is coming to q2 tomorrow, and the family will \textcolor{red}{celebrate her birthday} & Little Anh's back in District 2, Muzik! Let's \textcolor{teal}{carry her back home} tomorrow! & 57.53 & 80.30 \\
\bottomrule
\end{tabular}
\caption{Qualitative examples of translation errors by \texttt{Llama3.1-8B (Augmented)} that are overlooked by reference-free xCOMET but identified when using human references. The table shows the source text, model hypothesis, human reference, and the corresponding xCOMET scores with and without references.}
\label{tab:translation_examples}
\end{table*}

Overall, these patterns indicate benefits beyond lexical accuracy: \emph{Augmented} models better preserve pragmatic intent, select idiomatic equivalents, maintain fluency across switch points, and use acronym context to disambiguate meaning. Because lexically “correct” translations are not always the most useful or natural, the pipeline encourages learning latent rules of code-mixing and offers a practical recipe for other low-resource pairs where realism and quality control are essential.

\section{Discussion}\label{sec:discussion}


Our work presents a validated data augmentation methodology and a foundational resource for a challenging, low-resource code-mixed translation task. This section synthesizes key findings and their implications for multilingual NLP.

\subsection{Quality Trumps Quantity: Filtering is Crucial for Code-Mixed Augmentation}

For complex linguistic phenomena such as code-mixing, data quality is paramount. Fine-tuning on our \textsc{VietMix} seed corpus yields substantial gains over zero-shot, indicating that state-of-the-art multilingual models, as-is, underperform on natural, in-domain code-mixing. This establishes that a small amount of high-quality, in-domain data is an essential first step toward meaningful improvement.

\begin{figure}[t]
    \centering
    \includegraphics[width=0.5\textwidth]{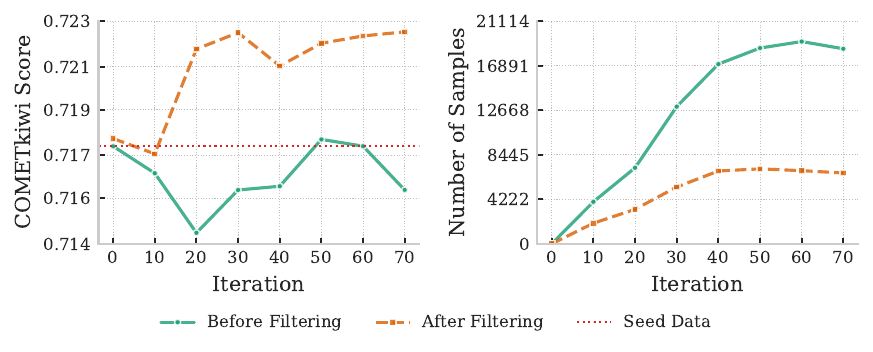}
    \caption{Impact of the Quality Assurance (QA) phase on synthetic data utility and yield across augmentation iterations. \textbf{(Left)} COMETkiwi scores on a development set for data before and after QA filtering, benchmarked against the seed-only data baseline (horizontal line). \textbf{(Right)} Number of synthetic samples generated per iteration before and after QA filtering, illustrating the yield of high-quality data.}
    \label{fig:analysis2}
\end{figure}

Beyond the seed effect, our augmentation experiments provide strong evidence for the quantity-quality trade-off. As shown in Figure~\ref{fig:analysis2}, adding unfiltered synthetic data depressed performance below the \emph{Seed-only} baseline, whereas quality-aware filtering restored and exceeded baseline. Thus, pruning low-quality synthetic examples is not optional but necessary. For specialized, low-resource MT, a smaller high-quality set is more valuable than a much larger noisy corpus.




\subsection{Human References Are Essential for Code-Mixed Evaluation}
Code-mixing challenges evaluations. While convenient, reference-free metrics that rely on source-hypothesis similarity can be misleading. To substantiate this claim, we re-evaluated \texttt{Llama3.1-8B} using both reference-based and reference-free variants of xCOMET. As shown in Table~\ref{tab:thread-xcomet}, reference-free scores were consistently artificially inflated, producing overconfidence that masks critical errors. Table~\ref{tab:translation_examples} provides qualitative snapshots of these failures. In the first example, the reference-free metric sees a superficial word overlap and gives this semantically unfaithful translation a score of $91.31$, while the reference-based metric assigns a much lower score of $71.43$. In the second, the model completely inverts the meaning of the source, yet the reference-free metric still assigns a higher score, $80.30$, due to surface-level alignment.

These observations underscore that evaluations of complex linguistic phenomena—such as code-mixing—must be anchored in human-crafted references. In this regard, \textsc{VietMix} supplies expert-curated, switch-sensitive references for reliable evaluation, enabling more faithful measurement of progress in real-world code-mixed usage.

\section{Conclusion}\label{sec:conclude}

This work introduces \textsc{VietMix}, an expert-translated benchmark for Vietnamese–English code-mixing that fills a key evaluation gap. We show that state-of-the-art models underperform on this task but improve markedly with the high-quality, in-domain data \textsc{VietMix} provides. We also present a quality-aware augmentation pipeline and show that principled filtering is essential, since unfiltered synthetic data degrades performance. Our method yields significant gains over strong baselines and produces translations that better capture pragmatic and idiomatic nuance. By releasing the benchmark and the pipeline, we offer a practical blueprint for addressing data scarcity and evaluation in other low-resource settings and enable more robust and culturally competent MT systems.


\section{Limitations}\label{sec:limitations}

We acknowledge several limitations. First, \textsc{VietMix} is sourced from Threads.com, and our augmentation data from Reddit and Bluesky. While this promotes stylistic consistency, linguistic patterns here may not generalize to other platforms, genres, or modalities. Future work should aim to create resources and evaluations that cover a broader range of platforms and genres. 

Second, our data pipeline relies on proprietary LLMs. This introduces limitations in reproducibility, cost, and the possibility of inheriting opaque biases. Therefore, a key avenue for future research is to use \textsc{VietMix} to distill the capabilities of these large models to smaller and open specialist models.

Third, our experiments exclusively evaluated the code-mixed to non-code-mixed \langcode{Vi}$\rightarrow$\langcode{En} direction. This deliberate choice targets our primary real-world application: enabling understanding of code-mixed text by monolingual speakers (e.g., in chat translation or content moderation). The inverse task of generating natural code-mixed from a non-code-mixed source (\langcode{En}$\rightarrow$\langcode{Vi}) is a fundamentally different challenge in stylization that goes beyond translation. Critically, this direction also presents immense evaluation challenges because a single non-code-mixed source can have an indefinite number of "correct" code-mixed targets, and this renders reference-based metrics inadequate. While linguistically fascinating, we have scoped this more ill-defined problem as future work. 

Fourth, while our expert annotation process was rigorous and accounted for dialectal diversity, its scale is necessarily limited. Expanding expert-curated corpora requires significant resources. Capturing the full spectrum of sociolinguistic variation remains an open challenge.

Finally, our evaluation centers on intrinsic translation quality. While we validated our metrics, we have not yet measured the impact of improved translation on more downstream, extrinsic tasks. This is an important next step for future work to assess the real-world utility of this methodology.

\section{Ethical Considerations}

Data making up the \textsc{VietMix} corpus was collected via a public API in accordance with the platform's terms of service. To protect user privacy, we implemented rigorous PII removal. We first applied an automated script to detect and delete posts with common PII patterns (i.e., phone numbers, email addresses). Second, every sample in the final data underwent manual review and anonymization to remove any remaining potentially sensitive information. As our work involved only publicly available data and professional translation services (not human subjects research), it did not require institutional review board (IRB) approval. \textsc{VietMix} (test set), all training and augmentation pipeline code will be publicly released under a CC BY-NC-SA 4.0 license to restrict use to non-commercial, research-only purposes.

To ensure linguistic inclusivity, translators and quality evaluators represent the three major dialectal variations of Vietnamese (Northern, Central, and Southern). Our annotation guidelines, developed collaboratively and refined through an adjudication process, will be released with the corpus to ensure transparency. We acknowledge that data sourced from social media platforms can reflect existing societal biases. While our sourcing from multiple platforms and quality filtering aim to mitigate the impact of potentially harmful content, models may still learn and reproduce subtle, harmful associations present in the training data. We consider the auditing of our models and corpus for social bias a critical area for future work.




\section{Acknowledgements}\label{sec:acknowledgement}

We thank Zoey Ki, Calvin Bao, and the members of the CLIP lab at University of Maryland for their constructive feedback. We also thank the annotators who contributed to building the test set. Their careful work was essential to this project.

\bibliography{anthology,custom}
\bibliographystyle{acl_natbib}

\clearpage
\appendix\label{sec:appendix}

\section{Data Augmentation Filtering} \label{appendix:filtering}
Our data augmentation consists of two stages with different mechanisms. This appendix section elucidates the methods employed in each.
\subsection{Heuristics}
In the first filtration stage, we implement heuristic constraints at the lexical and sub-lexical levels. These constraints eliminate pathological artifacts to make sentences more syntactically coherent.
\subsubsection{Proportional Length Equilibrium}
We compute the target-to-source token ratio and establish a permissible interval of [0.5, 1.5]. Candidate pairs exhibiting ratios outside this interval are immediately eliminated. This constraint mitigates catastrophic under-translation and verbose over-translation.
\subsubsection{Lexical Repetition}
To intercept repetition artifacts, we extract 5-word n-grams from the generated \langcode{Vi} translations and compute a weighted repetition coefficient:
\begin{equation}x
    R_{lex} = \frac{\sum_{g:count(g)>1} count(g)}{\sum_{g} count(g)}
\end{equation}
Candidate pairs with $R_{lex} \geq 0.3$ are eliminated because of lexical repetition redundancy. 
\newcolumntype{L}[1]{>{\centering\arraybackslash}p{#1}}
\begin{table*}[t]
    \centering
    \caption{Adapted MQM-Core evaluation criteria and rating scale (1-5, where 5 is best) used for human translator assessment and alignment discussions during the VietMix corpus creation. The table details the operationalization of six quality dimensions for evaluating code-mixed social media translations.}
    \renewcommand{\arraystretch}{1.3} 
    \small 
    \begin{tabular}{p{0.1\textwidth}|L{0.14\textwidth}|L{0.14\textwidth}|L{0.14\textwidth}|L{0.14\textwidth}|L{0.14\textwidth}}
    \toprule
        \textbf{Criterion} & \textbf{5} & \textbf{4} & \textbf{3} & \textbf{2} & \textbf{1} \\
    \midrule
        \textbf{Terminology} & Excellent handling of all terminology with appropriate decisions on cultural elements & Good handling of most terminology with minor inconsistencies & Some terms translated correctly, others inconsistently handled & Frequent terminology errors affecting understanding & Inappropriate handling of most cultural terms and stylistic choices \\
        \midrule
        \textbf{Accuracy} & Perfect semantic correspondence & Very good semantic alignment with minimal reorganization & Core meaning preserved but with some omissions/additions & Notable meaning loss or additions & Significant distortion of meaning \\
        \midrule
        \textbf{Linguistic Conventions} & Flawless grammar despite any source text errors & Good grammatical construction with very few errors & Acceptable grammar with some minor issues & Multiple grammatical issues affecting readability & Major grammatical errors making text difficult to understand \\
        \midrule
        \textbf{Style} & Excellent balance of faithfulness and natural expression & Natural-sounding with good interpretative choices & Balanced translation with some natural elements & Stiff translation with little interpretive flexibility & Extremely rigid, word-for-word translation \\
        \midrule
        \textbf{Locale Conventions} & Perfect handling of all locale-specific content and conventions & Good handling of locale conventions with appropriate notes & Adequate maintenance of most locale conventions & Poor handling of locale-specific content & Failed to maintain or explain any cultural elements \\
        \midrule
        \textbf{Audience Appropriateness} & Perfect social media tone maintained throughout & Good social media tone with minor variations & Generally appropriate with occasional inconsistencies & Frequently mismatched tone for the platform & Completely inappropriate tone for social media \\
    \bottomrule
    \end{tabular}
    \label{tab:MQM_eval}
\end{table*}
\begin{figure*}[h]
    \centering
    \begin{tabular}{|p{0.95\textwidth}|}
    \hline
    \\
\textbf{Your Main Task - Công việc chính:}
\begin{itemize}[leftmargin=1em, topsep=0pt, itemsep=0pt]
    \item You will see a series of Vietnamese sentences from social media - Anh/chị sẽ thấy một tập hợp câu được viết bằng Tiếng Việt từ mạng xã hội
    \item You will need to translate each sentence into natural, fluent English, or rate the quality of provided translations - Anh/chị sẽ cần dịch câu qua tiếng Anh với giọng văn chuẩn và tự nhiên, hoặc đánh giá chất lượng bản dịch được cho.
    \item We will ask you to capture both the meaning and tone of the original - Anh/chị sẽ cần giữ bản dịch gần nghĩa và hành văn của câu gốc nhất có thể.
\end{itemize}
\\
\textbf{Additional Guidelines - Chỉ dẫn thêm:}
\begin{itemize}[leftmargin=1em, topsep=0pt, itemsep=0pt]
    \item Depending on the question, you may type directly into the text box provided or select multiple-choice responses - Dựa trên yêu cầu của câu hỏi, anh/chị có thể gõ bản dịch vào ô `text' của bản khảo sát hoặc chọn phương án trắc nghiệm.
    \item You must translate the complete sentence and not skip any parts - Anh/chị sẽ cần dịch toàn bộ câu và không bỏ qua thành phần câu nào.
    \item You must maintain the original meaning and context as accurately as possible - Anh/chị sẽ cần giữ nghĩa và ngữ cảnh câu gần nhất có thể.
    \item You must use natural English expressions rather than word-for-word translations - Anh/chị sẽ cần tạo bản dịch Tiếng Anh tự nhiên nhất có thể, tránh dịch từng từ `word-for-word' kém tự nhiên.
    \item If there are slang terms or cultural references you recognize, and no appropriate English equivalents exist, you should simply describe it with a phrase - Trong trường hợp anh/chị gặp tiếng lóng `slang' hoặc dẫn chứng văn hoá Việt Nam khó dịch, và anh/chị không thể tìm bản dịch tiếng Anh phù hợp, ta sẽ dùng một cụm từ để mô tả từ đó.
\textit{For example - Ví dụ:} ``Nón quai thao = Vietnamese traditional headwear with a flat palm and fringes''; ``Đàn bầu = traditional Vietnamese instrument with one string and a cylindrical bamboo resonator''

    \item If there is Vietnamese-English code-mixing, you should retain English word forms in your translation if it does not alter meanings. - Trong trường hợp anh/chị gặp câu lẫn tiếng Anh-Việt, ta cần giữ cấu trúc của từ hoặc cụm từ tiếng Anh trong bản dịch nếu ngữ nghĩa không thay đổi.
    \item If there is dialectal Vietnamese word or grammatical structures you don't understand, you can skip that translation. - Trong trường hợp anh/chị gặp tiếng Việt địa phương mình không hiểu, vui lòng bỏ qua không dịch câu đó.
\end{itemize}
\\
\textbf{Important Notes - Yêu cầu quan trọng:}
\begin{itemize}[leftmargin=1em, topsep=0pt, itemsep=0pt]
    \item You can NOT to use online translation tools (like Google Translate) or AI assistants (like ChatGPT) - Anh/chị sẽ KHÔNG được sử dụng các công cụ dịch thuật online (như Google Translate) hoặc công cụ trí tuệ nhân tạo (như ChatGPT)
    \item You can NOT leave the task interface to look up concepts, as your authentic understanding suffices - Anh/chị sẽ KHÔNG được rời khỏi trang thực hành tác vụ để tra cứu từ, bản dịch chỉ cần anh/chị đưa ra ý hiểu tự nhiên nhất của bản thân.
    \item We value your authentic translation skills and cultural knowledge - Chúng tôi trân trọng công sức, kỹ năng dịch thuật tự nhiên, cũng như vốn hiểu biết văn hoá của anh/chị.
\end{itemize}
\\
    \hline
    \end{tabular}
    \caption{Bilingual (Vietnamese and English) instructions provided to professional translators. Key guidelines include maintaining meaning and tone, handling slang and cultural references, and retaining English word forms in code-mixed segments if meanings can be preserved.}
    \label{fig:task_instructions}
\end{figure*}
\subsubsection{Sub-lexical Repetition}
To intercept sub-lexical repetition artifacts, we extract 10-character n-grams, calculate their frequency distribution, and isolate the top  \(k=\min(\lfloor\sqrt{N}\rfloor,\,N-U)\) most frequent sequences, where $N$ is the total n-gram count and $U$ is the unique n-gram count. Then, we compute: 
\begin{equation}
    R_{char} = \frac{\sum_{i=1}^{k} f_i}{\sum_{i=1}^{n} f_i}
\end{equation}
Candidate pairs with $R_{lex} \geq 0.2$ are eliminated. This threshold and our threshold above for lexical repetition are manually determined based on qualitative observation of generated artifacts. 
\subsubsection{Code-mixing Equilibrium}
To preserve a relatively authentic code-mixing ratio, we implement a constraint that English tokens within the generated Vietnamese must not exceed 30\% of total tokens. We derive this parameter from an empirical analysis of natural code-mixing distributions in our real code-mixed corpus. This proportion represents the median value observed across our real Threads.net corpus, though variations exist depending on topic. 
\\
\\
All heuristic constraints function as first-stage filtration and eliminate approximately 37.8\% of generated synthetic pairs. 

\subsection{Neural Classifier}
Our second filtration stage seeks to approximate a naturalness assessment of generated texts. To approximate generated pairs' semantic-pragmatic naturalness, we implement a neural classification architecture described in the following subsections. 
\subsubsection{Model Selection and Classification}
We fine-tune XLM-RoBERTa,\footnote{https://huggingface.co/FacebookAI/xlm-roberta-base} given its effectiveness in cross-lingual tasks, to distinguish authentic Vietnamese code-mixed text from synthetic generations. The classifier received positive exemplars from realistic human-written text in our \langcode{Vi} Threads.net corpus and negative exemplars we generated from two sources:
\begin{itemize}
    \item Low-quality generations with lots of visible, unrealistic artifacts from Llama-3.2-3B,\footnote{https://huggingface.co/meta-llama/Llama-3.2-3B} selected for demonstrable deficiencies in \langcode{Vi} language modeling. 
    \item Moderate-quality but still containing many synthetic artifacts from Llama-3.1-8B.\footnote{https://huggingface.co/meta-llama/Llama-3.1-8B}
\end{itemize}
Both Llama models underwent initial fine-tuning on our seed dataset to translate \langcode{En} to \langcode{Vi}. We use these models to generate synthetic Vietnamese data when prompted with samples from \langcode{En} datasets described in \cref{sec:augmentation}. This stratified negative sampling helps the classifier develop sensitivity and recognize more patterns in synthetic text. 

\subsubsection{Implementation in Filtering}
After candidate \langcode{Vi} samples pass through the heuristic filters described in Section A.1, samples that satisfy heuristic constraints undergo neural classification. A straightforward decision rule is applied: samples receiving a "synthetic" classification confidence score of $\geq 0.5$ are discarded. This threshold is chosen to balance maintaining corpus size and ensuring some naturalness in generated texts. It is very challenging for synthetically generated code-mixed texts to achieve high levels of resemblance with human-written texts. Because of that, it is not our goal to invent a methodology for synthetic data to become as close as possible to the naturalness of human texts. This classification component only seeks to eliminate most semantically and pragmatically implausible code-mixed constructions that might otherwise contaminate our training data. 

\subsection{Filtering Approaches Evaluated but Not Incorporated in our Final Pipeline Implementation}
\subsubsection{Back-translation}
Back-translation, despite being a well-established technique for monolingual data augmentation, presents challenges for our scenario. Preliminary experiments with back-translation reveal significant distortions in code-mixing ratios. The back-translation model frequently "corrected" code-mixing by normalizing to either predominantly \langcode{Vi} or \langcode{En}. Back-translation also compounds errors at language switching points, per our qualitative examination. We believe that evaluating back-translated outputs in code-mixed scenarios, especially \langcode{Vi}, is especially challenging due to the variability in acceptable code-mixing patterns. This variability also creates challenges in establishing quality thresholds. 
\subsubsection{Direct Preference Optimization (DPO)}
Members who are native Vietnamese speakers in the research team together annotated for human preference. We performed comparative evaluations of multiple translation outputs for the same source text and assessed naturalness on a calibrated scale. Our results rendered this approach impractical for our current work due to subjective preference inconsistencies arising from regional dialectal variation. Annotators may disagree on "cleaned-up" code-mixing or the retention of messy code-mixing in real-world social media data. Preference-based optimization, therefore, was not performed to allocate efforts to developing our core filtering methodology. Future research could explore how to resolve inter-annotator disagreements in code-mixed language annotations arising from stylistic variations.    

\subsection{Language Identification Evaluation}\label{app:lid}

In the absence of a standard En--Vi code-mixed LID benchmark, we manually annotated a 400-sample set drawn from Bluesky to stress challenging phenomena (e.g., named entities, dates, orthographic overlap). Each sample was labeled as \emph{monolingual~Vi} or \emph{code-mixed}. We evaluated three approaches on this set: (i) \texttt{lingua} (dictionary/statistical), (ii) a single GPT-4o classifier, and (iii) a majority-vote \emph{Ensemble LLMs} (GPT-4o, Claude~3.5, Gemini~1.5). For reporting, we use standard binary metrics—Accuracy, Precision, Recall, and F1—treating \emph{code-mixed} as the positive class. Table~\ref{tab:lid-results} summarizes results.

The two-stage filter used in the main pipeline aligns with these results: an initial \texttt{lingua} pass efficiently removes clear monolingual cases, while the LLMs leverage context to disambiguate orthographically identical tokens and named entities. In the 400-sample set, error analysis indicates fewer than 5\% of cases involve named-entity misclassification (e.g., \emph{Vietnam} in an English context), supporting the reliability of code-mixed labels for downstream curation.

\begin{table}[t]
  \centering
  \caption{LID Evaluation Results}
  \label{tab:lid-results}
  \resizebox{\columnwidth}{!}{%
  \begin{tabular}{l S[table-format=2.2] S[table-format=2.2] S[table-format=2.2] S[table-format=2.2]}
    \toprule
    {Approach} & {Accuracy} & {Precision} & {Recall} & {F1 Score} \\
    \midrule
    Lingua          & 73.25 & 98.65 & 40.78 & 57.71 \\
    GPT-4o (single) & 88.00 & 83.25 & 91.62 & 87.23 \\
    Ensemble LLMs   & 89.75 & 92.59 & 83.80 & 87.98 \\
    \bottomrule
  \end{tabular}%
  }
\end{table}

\section{MQM-Core Dimensions Contextual Adaptions for Human Translators' Ratings and Alignment Discussions} \label{sec: MQM}

Table \ref{tab:MQM_eval} displays our adapted MQM-Core quality rating metrics descriptions. We operationalized six out of seven evaluative dimensions and created a Likert scale measurement framework for each dimension to facilitate translators' discussions throughout their annotation work and structure the team's collaborative discussions. Productive discussions on dimensions in which professional human translators most disagree with each other inform us where personal style and interpretive experience most influence translations. In this context, since innovative language use is common, maximizing naturalness in target \langcode{En} translations can mean sacrificing semantic equivalence. 

Contemporary neural evaluation systems predominantly align with technical translators' strict accuracy preferences. Hence, they largely overlook fluid interpretative ranges that non-technical translators rely upon to carry out and evaluate their work, especially in contexts abundant with cultural meanings. Future research may empirically examine how heterogeneity in translators' profiles (e.g., technical translators vs. literary translators) can inform the creation of evaluation metrics that dynamically adapt their criterion weights based on stylistic and domain heterogeneity.

Figure \ref{fig:task_instructions} snapshots the interface we show translators each session before they produce translations to ground their task interpretations on specific instructions.

\section{Experimental Details}\label{appendix:experimental_details}
All models were fine-tuned using Low-Rank Adaptation (LoRA) \cite{hu2021loralowrankadaptationlarge} to efficiently adapt pre-trained LMs to our \langcode{Vi}–\langcode{En} translation task. The fine-tuning configuration was:

\paragraph{Training Hyperparameters}
\begin{itemize}
    \item Learning rate: $2.5 \times 10^{-4}$
    \item Epochs: 3
    \item Batch size: 128
    \item Warmup steps: 10
    \item Precision: BF16
    \item LoRA rank: 256
    \item LoRA modules: \texttt{[q\_proj, k\_proj, v\_proj, o\_proj]}
\end{itemize}

All experiments and model evaluations reported in this paper were conducted on the held-out test set described in \cref{sec:vietmix}, after model development and hyperparameter tuning on the development set.

\paragraph{BLEU Results}

\begin{table}[t]
  \centering
  \tiny
  \setlength{\tabcolsep}{3pt}
  \resizebox{\columnwidth}{!}{%
    \begin{tabular}{lccc}
      \toprule
      \textbf{Setting} & \texttt{Qwen2.5-7B} & \texttt{Llama3.1-8B} & \texttt{GemmaX-9B} \\
      \midrule
      \textbf{Back-Translation (BT)} & 20.48 & 19.97 & 20.44 \\
      \textbf{Seed (9K)}             & 23.02 & 22.98 & 24.23 \\
      \textbf{Augmented}             & 22.40 & 22.14 & 22.71 \\
      \bottomrule
    \end{tabular}%
  }
  \caption{Detokenized BLEU on \langcode{Vi}$\rightarrow$\langcode{En} \textsc{VietMix} test.}
  \label{tab:bleu_appendix}
\end{table}

BLEU results are provided for comparability and are \emph{not} used as headline metrics (see main text). We report \textbf{detokenized BLEU} computed with \textsc{SacreBLEU} \cite{post-2018-call} using the default \texttt{13a} tokenizer and case-sensitive (mixed) scoring, with exponential smoothing. Exact scores appear in Table~\ref{tab:bleu_appendix}.

\section{Prompts}\label{appendix:prompts}
\subsection{Seed data generation} \label{appendix:seed_prompts}
Vietnamese code-mixed seed data underwent a two-stage refinement. Initially, we send samples from the seed data to GPT-4o, Gemini-1.5, and Claude-3.5-sonnet for detecting code-mixing, identifying Personally Identifiable Information (PII), and performing an initial translation; samples identified with PII were discarded. Subsequently, PII-cleared code-mixed samples were translated, primarily by GPT-4o. In cases of GPT-4o failure or poor output, translations were selected from Gemini or Claude. The prompt for this multi-model ensemble (also described in \cref{sec:augmentation}) is in Figure~\ref{fig:prompt}.
\begin{figure*}[h]
    \centering
    \small
    \begin{tcolorbox}[
    prompt,
    title={\textbf{Seed Data Detection and Translation}},
    ]
Detect if the given Vietnamese text contains any English words. If English words are present, detect if the given text contains any personally identifiable information (PII), then translate the entire text into English.\\
\\
\# Steps\\
0. *Language Detection*: Detect whether the given text contains English words.\\
1. *PII Detection*: Detect whether the text contains any PII (e.g. name, address, email, phone number, bank account).\\
2. *Translation*: Translate the entire text into English accurately.\\
\\
\# Output Format\\
- *Detection Result*: Output "True" if English words are present, otherwise "False".\\
- *PII Result*: "True" if any PII (e.g., name, address, email, phone number, bank account, etc.) is present, otherwise "False".\\
- *Translation*: If applicable, provide the English translation.\\
\\
\# Examples\\
*Example 1:*\\
\\
Input: "Tôi thích đi shopping vào cuối tuần."\\
- Detection Result: True\\
- PII Result: False\\
- Translation: "I like to go shopping on weekends."\\
\\
*Example 2:*\\
\\
Input: "Tôi thích đi mua sắm vào cuối tuần."\\
- Detection Result: False\\
\\
*Example 3:*\\
\\
Input: "Tôi tên là John, số điện thoại của tôi là 0987654321 và tôi thích đi shopping vào cuối tuần."\\
- Detection Result: True  \\
- PII Result: True\\
- Translation: "My name is John, my phone number is 0987654321, and I like to go shopping on weekends."\\
\\
\\
\# Notes\\
\\
- Only proceed to PII detection and translation if English words are detected.\\
- PII detection must detect correctly common identifiers (names, addresses, emails, phone numbers, bank accounts, etc.).\\
\\
$---$\\
\\
Input: \{input\}

    \end{tcolorbox}
    \caption{
    LLM prompt structure for the added data processing of naturally occurring Vietnamese-English code-mixed seed data. The prompt guides the model through a multi-step task: (1) detecting the presence of English words, (2) identifying any left-over personally identifiable information (PII) undetected in initial processing, and (3) translating the entire text into English if code-mixing is detected.
    \label{fig:prompt}}
\end{figure*}
\subsection{LLM-as-a-Judge translation assessments} \label{appendix:judge_prompts}
\begin{figure*}[t]
\centering
\begin{lstlisting}[language=Python,breaklines=true,showstringspaces=false,literate={í}{{\'i}}1]
class CompareEnglishTranslations(dspy.Signature):
    """You are an expert bilingual linguist in Vietnamese and English. Please choose the better English translation for the given Vietnamese source, using the provided reference translation to understand the intended meaning of the source. First, provide a rationale explaining your thought process, then
    determine which translation is better. Output exactly one of: A, B, Tie, or BothBad"""

    source = dspy.InputField(
        desc="The Vietnamese source text."
    )
    translation_a = dspy.InputField(
        desc="The English translated version from model A."
    )
    translation_b = dspy.InputField(
        desc="The English translated version from model B."
    )
    reference = dspy.InputField(
        desc="A human written English reference translation."
    )
    rationale = dspy.OutputField(
        desc="A brief explanation of the thought process."
    )
    better_translation: Literal['A','B','Tie','BothBad'] = dspy.OutputField(
        desc="Which translation is better? Output exactly one of: A, B, Tie, or BothBad."
    )
\end{lstlisting}
\caption{DSPy signature \texttt{(CompareEnglishTranslations)} defining the LLM-as-a-Judge prompt for evaluating translation quality. This prompt instructs an expert bilingual LLM to compare two English translations \texttt{(A and B)} of a Vietnamese source text against a human reference, provide a rationale, and determine the better translation \texttt{(A, B, Tie, or BothBad)}.}
  \label{fig:llmjudge_prompt}
\end{figure*}
Figure \ref{fig:llmjudge_prompt} presents the DSPy signature \texttt{(CompareEnglishTranslations)} defining the prompt used for the LLM-as-a-Judge evaluation of translation quality, as referenced in \cref{sec:experiments} and \cref{sec:results}.

\end{document}